\documentclass[10pt,twocolumn,letterpaper]{article}

\usepackage{iccv}
\usepackage{times}
\usepackage{epsfig}
\usepackage{graphicx}
\usepackage{amsmath}
\usepackage{amssymb}
\usepackage{booktabs}
\usepackage{tabularx}
\usepackage{float}
\usepackage{bbding}
\usepackage{setspace}

\usepackage[pagebackref=false,breaklinks=true,letterpaper=true,citecolor=black,colorlinks,linkcolor=black,bookmarks=false]{hyperref}

\iccvfinalcopy 

\def\iccvPaperID{636} 

\ificcvfinal\fi
\begin{document}

\title{Boundary-Aware Feature Propagation for Scene Segmentation}
\author{
Henghui Ding$^1$
\quad
Xudong Jiang$^1$
\quad
Ai Qun Liu$^1$
\quad
Nadia Magnenat Thalmann$^{1,2}$
\quad
Gang Wang$^3$\\
\vspace{-0.6em}
\\
$^1$Nanyang Technological University, Singapore\\
$^2$University of Geneva, Switzerland \qquad $^3$Alibaba Group, China
\\
{\tt\small \{ding0093, exdjiang, eaqliu, nadiathalmann\}@ntu.edu.sg, gangwang6@gmail.com}
}

\maketitle
\thispagestyle{empty}

\begin{abstract}
   In this work, we address the challenging issue of scene segmentation. To increase the feature similarity of the same object while keeping the feature discrimination of different objects, we explore to propagate information throughout the image under the control of objects' boundaries. To this end, we first propose to learn the boundary as an additional semantic class to enable the network to be aware of the boundary layout. Then, we propose unidirectional acyclic graphs (UAGs) to model the function of undirected cyclic graphs (UCGs), which structurize the image via building graphic pixel-by-pixel connections, in an efficient and effective way. Furthermore, we propose a boundary-aware feature propagation (BFP) module to harvest and propagate the local features within their regions isolated by the learned boundaries in the UAG-structured image. The proposed BFP is capable of splitting the feature propagation into a set of semantic groups via building strong connections among the same segment region but weak connections between different segment regions. Without bells and whistles, our approach achieves new state-of-the-art segmentation performance on three challenging semantic segmentation datasets,  i.e., PASCAL-Context, CamVid, and Cityscapes.
\end{abstract}

\vspace{-0.10cm}
\section{Introduction}
\vspace{-0.10cm}

\begin{figure}
\begin{center}
  \includegraphics[width = 0.486\textwidth]{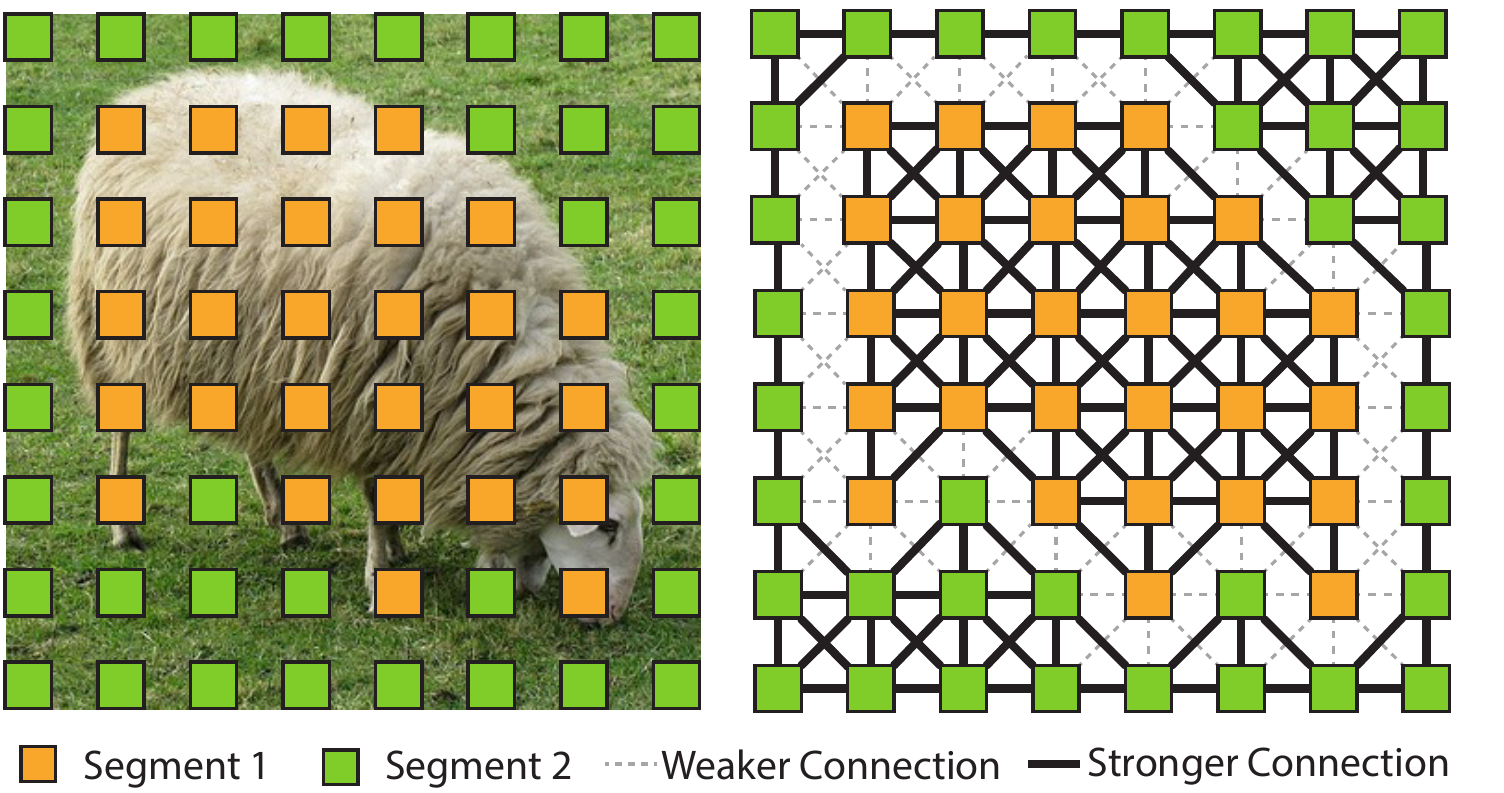}
\end{center}
\vspace{-0.24cm}
\caption{(Best viewed in color) The boundary-aware feature propagation module builds stronger connections within the same segment and weaker connections between different segments, which helps to enhance the similarity of features belonging to the same segment while keeping discrimination of features belonging to different segments.}
\vspace{-0.36cm}
\label{Figure:Introduction}
\end{figure}

Scene segmentation is a challenging and fundamental task that aims to assign semantic categories to every pixels of scene images. The key of scene segmentation refers to parsing and segmenting a scene image into a range of semantic coherent regions. Therefore, it is critical to improve the feature similarity of the same object while keeping the feature discrimination of different objects. To this end, on the one hand, we explore to propagate features throughout the images to share features and harvest context information, which is beneficial for improving the feature similarity. One the other hand, in order to keep the discriminative power of features belonging to different objects, we propose to make use of boundary information to control the information flow during propagation progress. In a word, we propose a boundary-aware feature propagation module to build strong connections within the same segment and weak connections between different segments, as shown in Figure \ref{Figure:Introduction}. This module requires two components: boundary detection and graph construction.


First, boundary detection, which is an implicit task in scene segmentation, is important for meticulous dense prediction. However, in existing segmentation methods, boundary detection did not attract due attention since boundary pixels only account for a small portion of the entire image and it has little contribution to the performance improvement. In this work, we try to find a way to simultaneously achieve segmentation and boundary detection, and further make use of the learned boundary information to enhance the segmentation performance. With regards to this, we propose to generate the boundary label of semantic objects from the existing object class labels given in the segmentation datasets and define it as an additional class for learning and classification. By doing so, concise boundaries are well learnt and inferred as one additional class because the characteristics of pixels on boundary are different from those of most pixels off boundary. And the parsing of pixels among disputed area (i.e., near the boundary) is enhanced. Moreover, taking boundary as an additional class requires little change on network but makes the network be aware of the boundary layout that could be further used for segmentation improvement.

Second, graphic model is needed to create the order rules for feature propagation. Convolutional methods~\cite{chen2016deeplab, yu2015multi} are popular in scene segmentation, but they usually consume large computation resources when aggregating features from grand range of receptive fields. Moreover, the convolution kernels could not vary with input resolutions, thus cannot ensure a holistic view of the overall image. Recently, DAG-RNN~\cite{shuai2017scene} proposes to use four directed acyclic graphs (DAGs) with different directions to model the function of undirected cyclic graphs (UCGs), which structurize the image by building pixel-by-pixel connections throughout whole image. However, DAGs require lots of loops to scan the images pixel by pixel. Thus it is very slowly even on low-resolution feature maps, which limits its application on ``dilated FCN''~\cite{chen2016deeplab, Zhao_2017_CVPR, Zhang_2018_CVPR} and on high resolution datasets like Cityscapes~\cite{cordts2016cityscapes}. To address this issue, we propose a more efficient graphic model to achieve faster feature propagation. We find that each DAGs adopted by~\cite{shuai2017scene} could be alternatively replaced by two Unidirectional Acyclic Graphs (UAGs), in which the pixels of the same row or column are dealt in parallel with 1D convolution. The proposed UAGs greatly speed up the feature propagation process. Moreover, different from the DAGs that are extremely deep, the proposed UAGs are much shallower and thus alleviate the problem of propagation vanish~\cite{pascanu2013difficulty}.

Finally, based on the UAG-structured image and the learned boundaries information, we build a boundary-aware feature propagation (BFP) module. In the BFP, local features of the same segment are shared via unimpeded connections to exchange information that achieves feature assimilation, while features of different segments are exchanged under controlled connections with the guidance of learned boundaries.
There are several advantages of our proposed boundary-aware feature propagation (BFP) network. First, since the proposed UAGs deal with pixels of the same row or column in parallel, we achieve the propagation process in a high speed. And the UAGs contain much fewer parameters than convolutional methods. Second, as we express boundary detection as classification of a semantic class, lots of parameters and complex module for boundary detection are saved. Third, with the advice of boundary confidence, the local features are propagated in a more motivated way, enhancing the similarity of features belonging to the same segment while keeping the discrimination of features belonging to different segments.

The main contributions of this paper can be summarized as follows:
\begin{itemize}
  \vspace{-0.26cm}
  \item We show that the boundary can be learned as one of semantic categories, which requires little change on network but obtains essential boundary information.
  \vspace{-0.26cm}
  \item We propose some unidirectional acyclic graphs (UAGs) to propagate information among high-resolution images with a high speed.
  \vspace{-0.26cm}
  \item We propose a boundary-aware feature propagation module to improve the similarity of local features belonging to the same segment region while keeping the discriminative power of features belonging to different segments.
  \vspace{-0.26cm}
  \item We achieve new state-of-the-art performance consistently on PASCAL-Context, CamVid, and Cityscapes.
  \vspace{-0.36cm}
\end{itemize}


\begin{figure*}
\begin{center}
  \includegraphics[width = 0.976\textwidth]{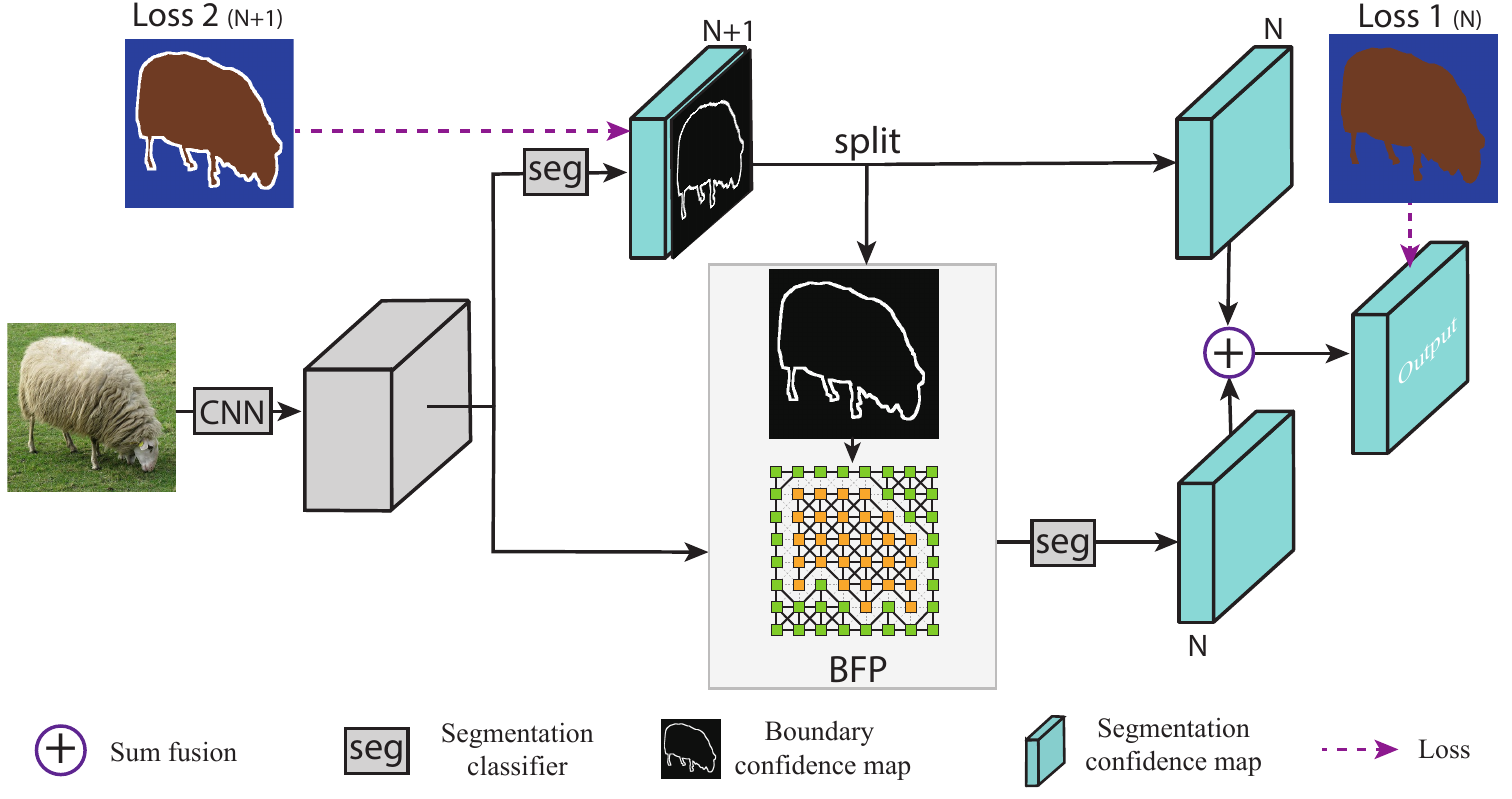}
\end{center}
\vspace{-0.36cm}
\caption{An overview of the proposed approach. We use the ResNet-101 (CNN)  with the dilated network strategy~\cite{chen2016deeplab} as backbone and the proposed boundary-aware feature propagation (BFP) module is placed on the top of CNN. The supervisor of loss 2 is the new ground truth of N+1 classes with an additional boundary class generated from the original ground truth of N classes.}
\vspace{-0.24cm}
\label{Figure:Framework}
\end{figure*}

\vspace{-0.10cm}
\section{Related work}
\label{Section:ReviewFCN}

\subsection{Scene Segmentation}
 Scene segmentation (or scene parsing, semantic segmentation) is one of the fundamental problems in computer vision and has drawn lots of attentions. Recently, thanks to the great success of Convolutional Neural Networks (CNNs) in computer vision~\cite{krizhevsky2012imagenet, szegedy2015going, Wang_ECCV2016ksr, liu2019feature, gu2018unpaired, wang2018detect, gu2019graph, ZengY_2018_CVPR, gu2019unpaired}, lots of CNNs based segmentation works have been proposed and have achieved great progress~\cite{he2019dynamic, fu2019adaptive, zeng2019joint, zhang2018bi, Zhang2019VOS, wang_iccv2019lfsd, rota2018place}. For example, Long et al.~\cite{long2015fully} introduce the fully convolutional networks (FCN) in which the fully connected layers in standard CNNs are transformed to convolutional layers. Noh et al.~\cite{noh2015learning} propose deconvolution networks to gradually upsample the coarse features to high resolution. Chen et al.~\cite{chen2016deeplab} propose to remove some pooling layers (or convolution stride) in CNNs and adopt dilated convolution to retain more spatial information. And some works focus on lightweight network architectures~\cite{badrinarayanan2017segnet, li2017not} and real-time segmentation~\cite{Zhao_2018_ICNet, paszke2016enet, Yu_2018_BiSeNet, romera2018erfnet}.

Context aggregation is a hot direction in scene segmentation. For example, Chen et al.~\cite{chen2016deeplab} propose an atrous spatial pyramid pooling (ASPP) module to aggregate multi-scale context information. Yu et al.~\cite{yu2015multi} employ multiple dilated convolution layers after score maps to exercise multi-scale context aggregation. Zhao et al.~\cite{Zhao_2017_CVPR} introduce pyramid spatial pooling~(PSP) to exploit context information from different scale regions.
Zhang et al.~\cite{Zhang_2018_CVPR} encode semantic context to network and stress class-dependent feature maps. He et al.~\cite{Hejun_2019_CVPR} propose adaptively pyramid context module to capture global-guided local affinity. Fu et al.~\cite{fu2019dual} integrate local and global dependencies with both spatial and channel attention. Ding et al.~\cite{ding2019semantic} employ semantic correlation to infer shape-variant context.

Graphic models have a long history in scene segmentation. Early works construct the graphic model with hand-crafted features~\cite{gould2009decomposing, liu2011sift, yang2014context, tighe2013finding}. Markov Random Fields (MRF)~\cite{gould2009decomposing, kumar2010efficiently, larlus2008combining} and Conditional Random Fields (CRF)~\cite{krahenbuhl2011efficient, russell2009associative, chen2016deeplab, liu2015semantic} build the dependencies according to the similarities of neighboring pixels. Liang et al.~\cite{liang2016semantic} propose to construct graph topology based on superpixel nodes and incorporate long-range context with Graph LSTM. Shuai et al.~\cite{shuai2017scene} adopt undirected cyclic graphs (UCGs) to formulate an image and decompose the UCGs with directed acyclic graphs (DAGs). Byeon et al.~\cite{Byeon_2015_CVPR} propose to divide an image into non-overlapping windows and employ a 2d LSTM to construct local and global dependencies. However, most of graph-based method are time-consuming and computationally expensive as they require candidate pre-segments, superpixels, or lots of loops.

In this work, we propose unidirectional acyclic graphs (UAGs), based on which the local features are quickly propagated in parallel. And to construct strong dependencies within the same segment and weak dependencies among different segments, we exploit the learned boundary information to guide the feature propagation within the UAG-structured image.

\subsection{Boundary Detection}
\vspace{-0.10cm}

Boundary detection is a fundamental component for many computer vision tasks and has a long history~\cite{arbelaez2011contour, dollar2013structured, konishi2003statistical, isola2014crisp}. For example, Lim et al.~\cite{lim2013sketch} propose sketch tokens (ST) and Doll{\'a}r~\cite{dollar2015fast} et al. propose structured edges (SE) based on fast random forests to deal with boundary detection as local classification problem. Recently the success of CNNs have great improve the performance of boundary detection~\cite{bertasius2015deepedge, bertasius2015high, hwang2015pixel, shen2015deepcontour, xie2015holistically}. Xie et al.~\cite{xie2015holistically} employ features from the middle layers of CNNs for boundary detection. Shen et al.~\cite{shen2017multi} propose multi-stage fully convolutional networks for boundary detection of electron microscopy images. These methods target at optimizing the accuracy of boundary detection instead of generating semantic boundary information for high-level tasks. Boundary information could be used for improving segmentation performance. For example, Bertasius et al.~\cite{bertasius2015high}, Hayder et al.~\cite{hayder2017boundary}, Chen et al.~\cite{chen2016semantic} and Kokkinos~\cite{kokkinos2015pushing} employ the binary edges to improve the segmentation performance. However, they all employ an additional branch of network for edge detection, which requires more resources and deal with segmentation and boundary detection as two detached tasks. Different from~\cite{bertasius2015high, hayder2017boundary, chen2016semantic}, our goal is not to detect the clearly binary boundaries, but to infer a boundary confidence map that represents the probability distributions of high-level boundary layout.


\section{Approach}
\vspace{-0.10cm}

Due to the diverse style and complex layout of scene images, it is necessary to classify every pixel using holistic context information but protect its differentiation from overwhelming by global scene. In this respect, we propose a boundary-aware feature propagation module to arm the local features with a holistic view of contextual awareness but keep the discriminative power of features for different objects. The overall architecture is shown in Figure \ref{Figure:Framework}. We use the dilated FCN (subsampled by 8) based on ResNet-101~\cite{he2016deep} as backbone. The supervisor of loss 2 is the boundary-aware ground truth (N+1 classes) that are generated from the original ground truth (N classes).

\subsection{Semantic Boundary Detection}
\vspace{-0.10cm}
Boundary delineation is favourable for meticulous scene parsing. However, because of the various semantic labels and complex layout of objects in segmentation datasets, parsing pixels in the boundary area is always difficult and results in confused prediction. In this work, instead of directly assigning semantic labels to pixels in boundary area, we explore to infer the boundary layout first and improve the segmentation performance with the learned boundary information. Lots of works have contributed to boundary detection~\cite{bertasius2015deepedge, bertasius2015high, hwang2015pixel, shen2015deepcontour, xie2015holistically}, but most of them focus on edges that sketch the objects. Different from them, we only focus on the boundaries of semantic objects that are predefined in segmentation datasets. We have observed that boundaries have the property of dramatic changes in RGB and feature information. And the boundary label is easy to be generated from the existing ground truth. Consequently, we assume that the boundary could be viewed as an additional semantic category and simultaneously learned with other existing object categories.
As shown in Figure~\ref{Figure:Framework}, we obtain a new ground truth (Loss 2, N+1 classes) from the original ground truth (Loss 1, N classes) and utilize the new ground truth for supervising the network to learn and infer the boundary layout. 

Different from previous boundary detection works that aim to boundary delineation only or deal with segmentation and boundary detection as two detached tasks, our proposed semantic boundary detection is embed with semantic object parsing. Our boundary detection module only targets at boundaries of semantic objects predefined in the training data and generate concise boundary information under interaction with segmentation. These two tasks are combined into one, and they benefit each other. By training them together, the scene segmentation classes help suppress the edges within objects that are not semantic boundary of objects, e.g., edges of eyes. Scene segmentation helps the boundary detection to filter out noise and delineate well-directed boundary, while boundary detection makes the scene segmentation be aware of the important boundary layout information.

\begin{figure}[t]
\begin{center}
  \includegraphics[width = 0.5\textwidth]{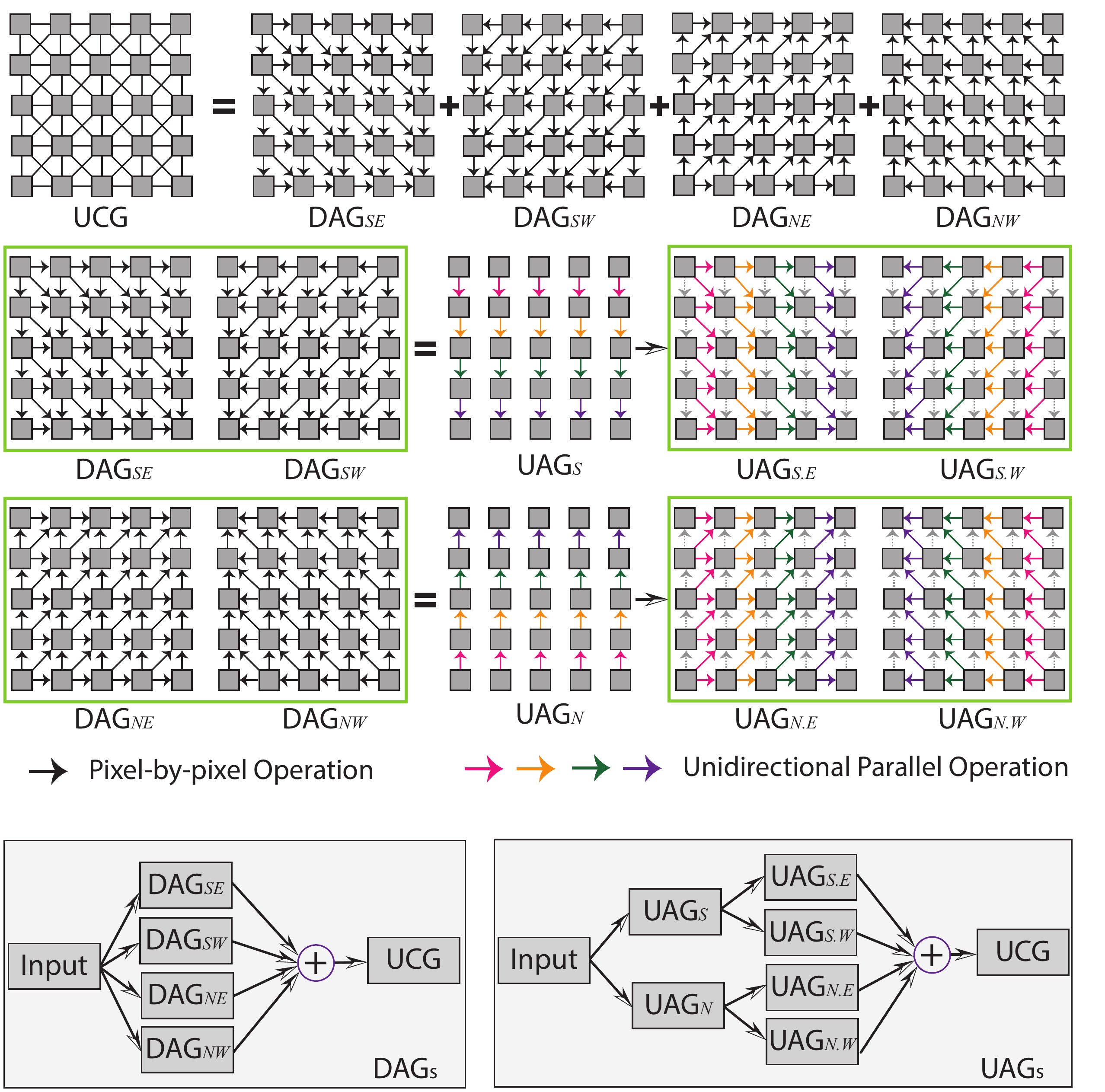}
\end{center}
\caption{Each point of the DAGs has three different directions. Thus, the DAGs have to scan the image pixel by pixel and consumes lots of time with many loops. We decompose the four DAGs to six Unidirectional Acyclic Graphs (UAGs). Each of UAGs propagates information toward a single direction, which deals with pixels of each row in parallel and then deals with pixels of each column in parallel. For example, the UAG${_S}$ is in the south direction only and the UAG$_{{S.E}}$ is in the east direction only (based on UAG${_S}$).}
\label{Figure:UCG_DAG}
\end{figure}

\begin{figure*}
\begin{center}
  \includegraphics[width = 0.996\textwidth]{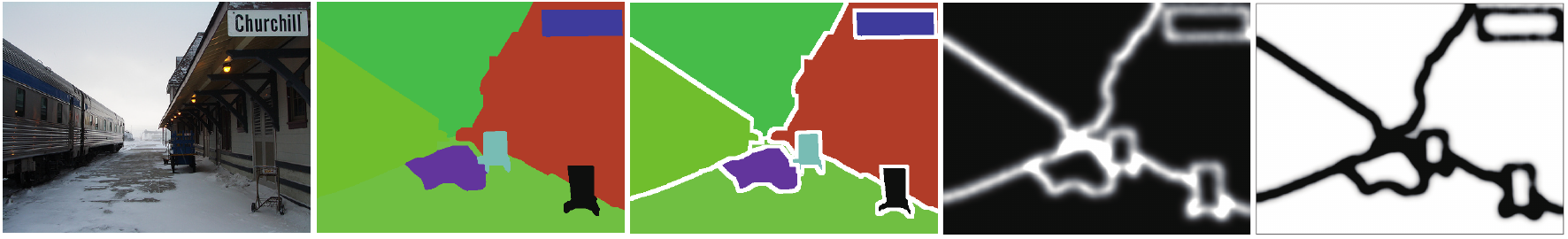}

  \scriptsize{\hspace{5em} \textbf{(a)}  \hfill \hspace{0em} \textbf{(b)} \hfill \hspace{0em} \textbf{(c)} \hfill \hspace{0em} \textbf{(d)} \hfill \hspace{0em} \textbf{(e)} \hspace{5em} } \\
\end{center}
\vspace{-0.36cm}
\caption{(a) Image; (b) Original ground truth; (c) New generated ground truth: add a boundary class generated from the original ground truth; (d) Boundary confidence map: the probability distribution of the boundary layout; (e) Propagation confidence map: the confidence distribution of propagation.}
\vspace{-0.24cm}
\label{Figure:ImageExample}
\end{figure*}

\subsection{Unidirectional Acyclic Graphs}

Context is designed to aggregate wide range of surrounding information, thus it desires a holistic view of the overall image without regard to resolution. One popular way is to employ stacked convolution or dilated convolution to enlarge the receptive field~\cite{chen2016deeplab, yu2015multi, Ding_2018_CVPR}, but this consumes large computation resources. The work of~\cite{shuai2017scene} proposes DAG-RNN to capture long-range context based on directed acyclic graphs (DAGs). As shown in Figure \ref{Figure:UCG_DAG}, pixels are locally connected to form a undirected cyclic graph (UCG) to build propagation channels among the whole image. To overcome the loopy property of UCG, the UCG is decomposed to four DAGs with different directions (southeast, southwest, northeast, northwest). However, since each pixel of the DAGs has three different directed connections, the feature propagation based on DAG-structured images has to scan the image pixel by pixel and requires lots of loops. Thus it is very slowly even on low-resolution feature maps, which limits its application on ``dilated FCN''~\cite{chen2016deeplab, Zhao_2017_CVPR, Zhang_2018_CVPR} and on high resolution datasets like cityscapes~\cite{cordts2016cityscapes} and CamVid~\cite{brostow2008segmentation}. To address this issue, we explore to reduce the number of loops and propagate information in parallel. Herein, we propose some Unidirectional Acyclic Graphs (UAGs) as shown in Figure \ref{Figure:UCG_DAG}, which deal with pixels of each row in parallel and then deals with pixels of each column in parallel. Each DAGs adopted by~\cite{shuai2017scene} could be alternatively replaced by two UAGs. For example, the DAG$_{SE}$ is decomposed to UAG$_{S}$ and UAG$_{S.E}$, where UAG$_{S}$ is south directed that deals with pixels of the same row in parallel and UAG$_{S.E}$ is east directed (after UAG$_{S}$) that deals with pixels of the same column in parallel. As a result, the number of loops for each DAGs is reduced from \emph{H$\times$W} to \emph{H$+$W}, where $H$ and $W$ are the height and width of feature maps. The proposed UAGs greatly speed up the feature propagation process, which is economic and desired in practice, especially for applications that require high resolution and big eyeshot (e.g., self-driving vehicle). Moreover, due to the pixel-by-pixel operation, the recursive layers in DAGs are very deep that could be unrolled to thousand layers. This causes the problem of propagation vanish~\cite{pascanu2013difficulty}. The proposed UAGs are much shallower than DAGs and thus alleviate the the problem of propagation vanish.



\vspace{-0.10cm}
\subsection{Boundary-Aware Feature Propagation}
\vspace{-0.10cm}


 However, unselective propagation would make the features assimilated, which results in smooth representation and weakens the features' discrimination. To classify features in different objects and stuff in scene segmentation, it is beneficial to improve the feature similarity of the same object while keeping the feature discrimination of different objects. Therefore, we introduce the boundary information into feature propagation to control the information flow between different segments. As shown in Figure \ref{Figure:Introduction}, with the learned boundary information, we build strong connections for pixels belonging to the same segment but weak connections for different segments. During propagation, more information is passed via strong connections within the same segment and less information flows crossing different segments. In such a way, pixels get more information from other pixels of the same objects and less information from pixels of other objects. Consequently, features of different objects could keep their discrimination while features of the same object trend to be cognate, which is desired for segmentation. The detailed process of proposed boundary-aware feature propagation is presented below.

\begin{figure}
\begin{center}
  \includegraphics[width = 0.396\textwidth]{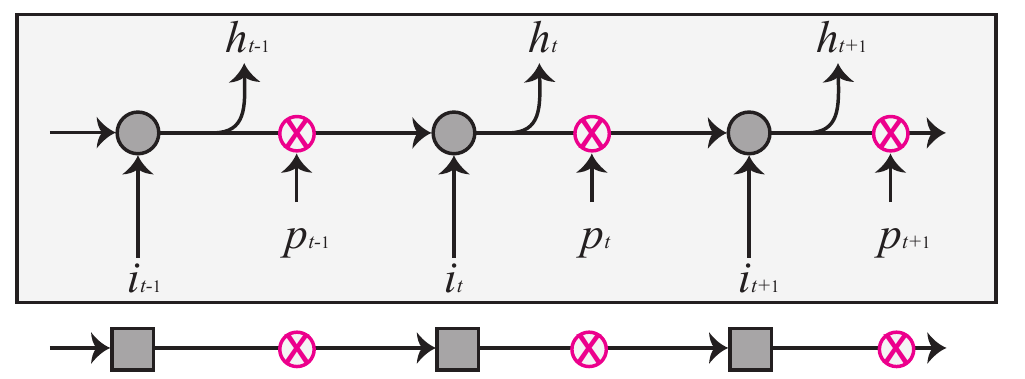}
\end{center}
\caption{As our UAGs are unidirectional and in parallel, we show the propagation process in 1D here for clarity. $i_{t}$ represents the feature of pixel at position $t$, $h_{t}$ is output (hidden state), $p_{t}$ is the propagation confidence.}
\label{Figure:Boundary-Aware-DAG}
\end{figure}

 As our UAGs are unidirectional and in parallel, we formulate the propagation process in 1D here for notation clarity. Extension to 2D/3D is straightforward. We denote the feature of pixel at position $t$ as $i_{t}$, and the corresponding output (hidden state) is denoted as $h_{t}$. The standard propagation process based on our UAG-structured image is formulated as following:
\vspace{-0.07cm}
 \begin{equation}
\begin{aligned}
& h_{t} = g(U*i_{t}+W*h_{t-1}+\delta)\\
\end{aligned}
\end{equation}
\vspace{-0.07cm}
where $*$ is 1D convolution operation, $U, W$ are learnable parameters of 1D convolution and $\delta$ is learnable bias. $g$ is element-wise nonlinear activation function (we use ReLU).

For the boundary-aware propagation, we first extract the boundary confidence map from the $(N+1)$ segmentation confidence maps, as shown in Figure \ref{Figure:Framework}. We denote the boundary confidence of  pixel $t$ as $b_{t}$, corresponding to $i_{t}$. Based on the boundary confidence map, we generate the propagation confidence map:
\vspace{-0.07cm}
\begin{equation}
p_{t} = 1-\beta f(\alpha b_{t}-\gamma)
\end{equation}
\vspace{-0.07cm}
where $p_{t}$ is the propagation confidence that decides how much the information of pixel $t$ to be passed to next region. $\alpha$=$20$ and $\gamma$=$4$ are constant chosen by experience, $f$ is sigmoid function to enhance the boundary, $\beta$ is a learnable parameter. With the propagation confidence, the propagation process can be reformulated as below:
\vspace{-0.07cm}
\begin{equation}
\begin{aligned}
& h_{t} = g(U*i_{t}+W*h_{t-1}p_{t-1}+\delta)\\
\end{aligned}
\end{equation}
\vspace{-0.07cm}
as shown in Figure~\ref{Figure:Boundary-Aware-DAG}, the propagation is controlled by the boundary and thus models boundary-aware context feature for better parsing of different segments.

For the UAGs that have ``two directions'', they are also unidirectional and in parallel. For example, UAG$_{S.E}$ is formulated below:
\vspace{-0.07cm}
\begin{equation}
h_{t}^j = g(U*i_{t}^j+W*h_{t-1}^jp_{t-1}^j+ \hat{W}*h_{t-1}^{j-1}p_{t-1}^{j-1} +\delta)
\end{equation}
\vspace{-0.07cm}
There are two hidden states, $h_{t-1}^j$ and $h_{t-1}^{j-1}$, input to the current cell, where $t$ and $j$ are the denotations of horizontal and vertical axis, respectively. Finally, the hidden states of the corresponding positions in four UAGs (i.e., UAG$_{S.E}$, UAG$_{S.W}$, UAG$_{N.E}$, UAG$_{N.W}$) are fused together to generate the final output, as shown in Figure \ref{Figure:UCG_DAG}.

One example of boundary confidence map and propagation confidence map is shown in Figure \ref{Figure:ImageExample}. We learn the boundary confidence map under the supervise of new generated ground truth with additional boundary class. To control the progress of feature propagation, propagation confidence map is generated from boundary confidence map. If pixel $i_{t}$ is in boundary region, then it has a higher boundary probability $b_{t}$ and hence a smaller propagation probability $p_{t}$. Thus, the feature propagation is suppressed and weak signals are passed to next pixel. Otherwise, it has a strong propagation to spread its features to next pixel.

\section{Experiments}

\subsection{Implementation Details}
\label{Section:implementation_detail}

Our Network is implemented based on the public platform Pytorch. We use ResNet-101~\cite{he2016deep} with the dilated network strategy~\cite{chen2016deeplab} (subsampled by 8) as our backbone.
In detail, the convolution strides for downsampling in last two blocks are reset to 1 and the convolutions of last two blocks are dilated with dilation rates of 2 and 4 receptively. $Pool~5$ and layers after it are discarded.
The network is trained with mini-batch, batchsize is set to 12 for PASCAL-Context and 8 for Cityscapes and CamVid.
Following deeplab-v2~\cite{chen2016deeplab}, we use the ``poly'' learning rate policy, where current learning rate $ Lr_{c} $ depends on the base learning rate $ Lr_{b} $ and iterations: $Lr_{c}=Lr_{b}$$\times$$(1-\frac{iter}{total\_iter})^{0.9}$. Momentum and weight decay are fixed to 0.9 and 0.0001 respectively.
~We adopt random horizontal flipping and random resizing between 0.5 and 2 to augment the training data.

Most of the scene segmentation datasets do not provide boundary layout, we use the provided segmentation ground truth to generate boundary-aware ground truth, as shown in Figure \ref{Figure:ImageExample} (c). As we adopt the dilated FCN as our backbone, the spatial size is downsampled 8 times in encoding process. Thus to avoid the loss of boundary information in feature maps with smallest spatial size, pixels with distance smaller than 9 pixels (i.e., trimap of 18 pixels) to boundary are defined as boundary pixels and their ground truth labels are set to $N+1$, where N is the number of classes in datasets. In our experiments, the over-wide boundary (e.g., trimap of 50 pixels) squeezes small objects and weakens the function of boundary in feature propagation. We evaluate our network with mean Intersection-over-Union (\textbf{mIoU}). Mathematical definition of mIoU please refer to~\cite{long2015fully}.

\subsection{Efficiency Analysis}

To evaluate the speed of the proposed UAGs, we report in Table~\ref{Table:DAGvsUAG} the inference time of UAGs and compared it with DAGs~\cite{shuai2017scene} on different resolution of input images, based on dilated FCN (subsampled by 8). The number of loops is also recorded. Different from DAGs that have to scan the image pixel by pixel, the proposed UAGs deal with pixels of each row/column in parallel, hence they save a lot of loops. As shown in Table \ref{Table:DAGvsUAG}, the UAGs contain much less loops than DAGs and consequently they are much faster than DAGs. Especially with high resolution (e.g., 960$\times$720), DAGs are very slow and time-consuming. The speed of DAGs is highly related with the input resolution that determines the number of loops, thus DAGs are not suitable of high-resolution datasets (e.g., Cityscapes~\cite{cordts2016cityscapes}) and FCN with dilated network strategy~\cite{chen2016deeplab}. Besides the inference time, training of DAGs based on dilated FCN requires one hundred times more GPU hours than our proposed UAGs, which also shows the high efficiency of our approach. To quantitatively compare the segmentation performances of DAGs and proposed UAGs, we evaluate them on VGG-16~\cite{simonyan2014very} with encoder-decoder strategy, exactly in the same way like that in~\cite{shuai2017scene}. DAGs and UAGs achieve almost the same results on PASCAL-Context (UAGs 43.0\% vs.~DAGs 42.6\%). This shows that the proposed UAGs realize the same function as DAGs but with a much faster speed.

\begin{table}[t]
\footnotesize
\begin{center}
\begin{tabular}{lccc}
\toprule
Methods & Input Resolution&\# Loops & Time (s)\\
\midrule
\midrule
FCN& $480\times 360$ &none &0.35\\
DAGs& $480\times 360$&10800 &17.92\\
UAGs& $480\times 360$&300 &0.47\\
\midrule
FCN& $960\times 720$&none &0.42\\
DAGs& $960\times 720$&43200 &56.97\\
UAGs& $960\times 720$&600 &0.76\\
\bottomrule
\end{tabular}
\end{center}
\vspace{-0.24cm}
\caption{Inference speed comparison of FCN~(baseline), DAGs, and UAGs on dilated ResNet-101 with different resolution inputs.}
\vspace{-0.24cm}
\label{Table:DAGvsUAG}
\end{table}
\subsection{Ablation Studies}

We show the detailed results of ablation studies of the proposed approach in Table~\ref{Table:AblationStudy}. The proposed UAGs harvests local features and propagate them throughout the whole image to achieve a holistic context aggregation, which greatly improves the segmentation performance from the baseline (dilated FCN). Then, based on the UAGs, we learn the boundary information and inject it to the propagation process to control the information flow between different regions. With the boundary information, the UAGs build stronger connections of pixels within the same segment and weaker connections between different segments. Thus, features of the same segment become more similar but features of different segments remain discriminative.

\begin{figure}[t]
\centering
  \scriptsize{\hspace{4em} Images  \hfill \hspace{1em} Boundary Confidence \hfill  Ground Truth\hspace{1.9em} } \\
 \includegraphics[width=0.487\textwidth]{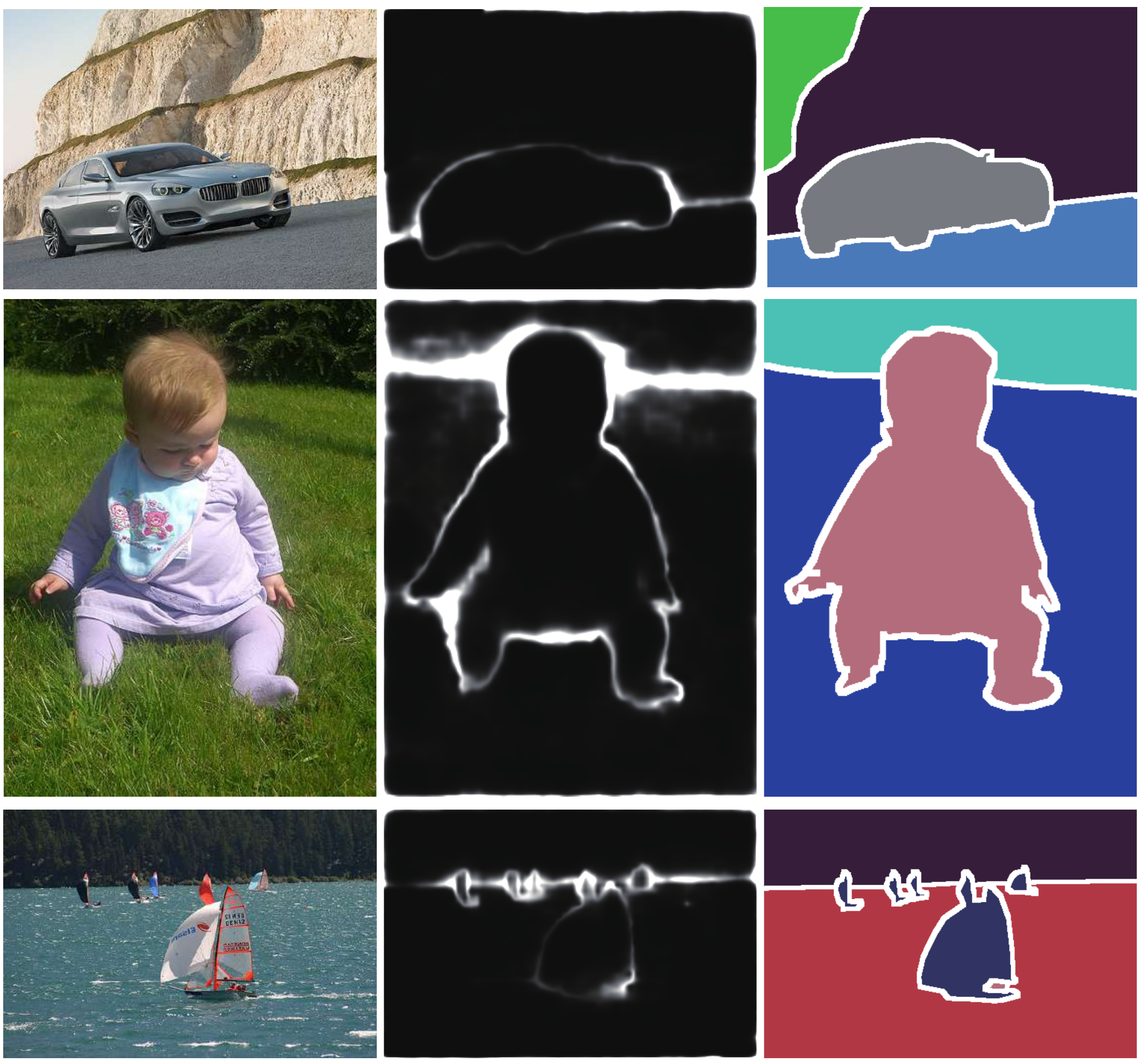}
\vspace{-0.24cm}
\caption{Qualitative examples of inferred boundary map.}
\vspace{-0.26cm}
\label{Figure:Boundary_demos}
\end{figure}

\begin{figure}[t]
\begin{center}
  \includegraphics[width = 0.266\textwidth]{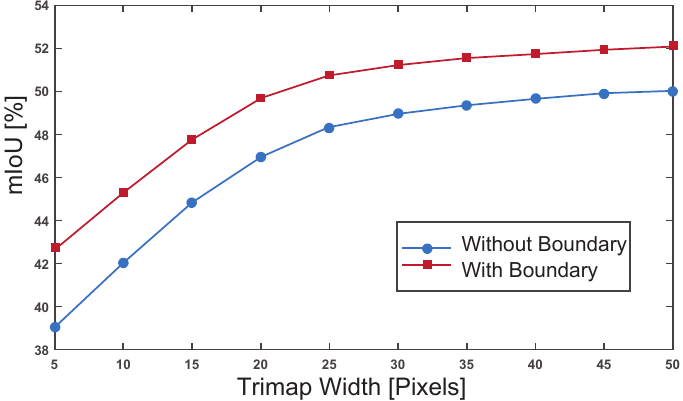}
\end{center}
\vspace{-0.36cm}
\caption{Segmentation performance within band (trimap) around boundaries.}
\vspace{-0.16cm}
\label{Figure:trimap}
\end{figure}

We also visualize some examples of the inferred boundary confidence map in Figure \ref{Figure:Boundary_demos}. As shown in Figure \ref{Figure:Boundary_demos}, the inferred boundary map mainly involves boundaries between the semantic segments predefined in datasets instead of other edge information. Therefore, it contains the desired boundary layout of semantic objects and could be used for control of the feature propagation throughout image. The results in Table~\ref{Table:AblationStudy} show the effectiveness of the boundary-aware feature propagation (BFP) network. Following~\cite{kohli2009robust, chen2016semantic}, we evaluate the performance of BFP near boundaries, as shown in Figure~\ref{Figure:trimap}. We compute the mIoU for the regions within different bands of boundaries.

DT-EdgeNet~\cite{chen2016semantic} is the most related to BFP. However, BFP learns the boundary as one of semantic classes and is trained end-to-end, while DT-EdgeNet learns edge with additional EdgeNet and requires a two-step training process for DeepLab and EdgeNet. As shown in Figure~\ref{Figure:Boundary_demos}, Our learned boundaries response less to object interior edges than DT-EdgeNet.
BFP is proposed to perform feature propagation that is some kind of contextual feature modeling, while DT is used to refine the segmentation scores. We use the DT to filter the segmentation scores of BFP and this brings 0.7\% performance gain, which shows that the DT and BFP are complementary.

\begin{table}[t]
\footnotesize
\begin{center}
\begin{tabular}{lccccc}
\toprule
Methods&Backbone & UAGs& Boundary & MS &mIoU \\
\midrule
\midrule
FCN & ResNet-50& & & &41.2\\
BFP & ResNet-50 & \Checkmark& & &49.8\\
BFP & ResNet-101 & \Checkmark& & &50.8\\
BFP & ResNet-101 & \Checkmark& \Checkmark& &52.8\\
BFP & ResNet-101& \Checkmark& \Checkmark& \Checkmark&53.6\\
\bottomrule
\end{tabular}
\end{center}
\vspace{-0.16cm}
\caption{Ablation Study of Boundary-aware Feature Propagation (BFP) Network on PASCAL-Context. Baseline is dilated FCN and MS means multi-scale testing.}
\vspace{-0.1cm}
\label{Table:AblationStudy}
\end{table}

\subsection{Comparison with the State-of-the-Art Works}
\label{Section:BenchMarkResults}

\textbf{PASCAL-Context}~\cite{mottaghi_cvpr14} provides pixel-wise segmentation annotation for the whole scene image. There are 4998 training images and 5105 testing images in PASCAL-Context. Following~\cite{mottaghi_cvpr14}, we use the most common 59 classes for evaluation. Testing accuracies of PASCAL-Context are shown in Table \ref{Table:resultsContext}, which shows that the proposed BFP outperforms the state-of-the-art works by a large margin.

\begin{table}[t]
\footnotesize
\begin{center}
\begin{tabular}{p{3.76cm}|p{1.276cm}<{\centering}}
\toprule
Methods & mIoU \\
\midrule
\midrule
O2P~\cite{carreira2012semantic}& 18.1 \\
FCN-8s~\cite{shelhamer2016fully} & 39.1 \\
BoxSup~\cite{dai2015boxsup} & 40.5\\
HO-CRF~\cite{arnab2016higher}& 41.3 \\
PixelNet~\cite{bansal2016pixelnet} & 41.4 \\
DAG-RNN~\cite{shuai2017scene} & 43.7 \\
EFCN~\cite{shuai2019toward}& 45.0 \\
DeepLab-v2+CRF~\cite{chen2016deeplab} & 45.7 \\
RefineNet~\cite{Lin_2017_CVPR}& 47.3 \\
MSCI~\cite{lin2018multi} & 50.3 \\
CCL+GMA~\cite{Ding_2018_CVPR} & 51.6 \\
EncNet~\cite{Zhang_2018_CVPR} & 51.7 \\
\midrule
BFP (ours) & \textbf{53.6} \\
\bottomrule
\end{tabular}
\end{center}
\vspace{-0.24cm}
\caption{Testing accuracies on PASCAL-Context.}
\vspace{-0.02cm}
\label{Table:resultsContext}
\end{table}

\begin{table}[t]
\footnotesize
\begin{center}
\begin{tabular}{p{3.76cm}|p{1.276cm}<{\centering}}
\toprule
Methods & mIoU \\
\midrule
\midrule
DeconvNet~\cite{noh2015learning} &48.9  \\
SegNet~\cite{badrinarayanan2017segnet}  &50.2 \\
FCN-8s~\cite{long2015fully}  &52.0 \\
DeepLab~\cite{chen2015semantic}  &54.7 \\
DilatedNet~\cite{yu2015multi} &65.3 \\
Dilation+FSO~\cite{kundu2016feature}&66.1 \\
G-FRNet~\cite{islam2017gated} &68.0\\
Dense-Decoder~\cite{Bilinski_2018_CVPR} &70.9\\
\midrule
BFP (ours) &\textbf{74.1}  \\
\bottomrule
\end{tabular}
\end{center}
\vspace{-0.2cm}
\caption{Testing accuracies on CamVid.}
\vspace{-0.36cm}
\label{Table:resultsCamVid}
\end{table}

\begin{table*}[t]
\footnotesize
\begin{center}
\begin{tabular}{p{1.8cm}|p{0.3cm}<{\centering}p{0.3cm}<{\centering}p{0.3cm}<{\centering}p{0.3cm}<{\centering}p{0.3cm}<{\centering}p{0.3cm}<{\centering}p{0.3cm}<{\centering}
p{0.3cm}<{\centering}p{0.3cm}<{\centering}p{0.3cm}<{\centering}p{0.3cm}<{\centering}p{0.3cm}<{\centering}p{0.3cm}<{\centering}p{0.3cm}<{\centering}p{0.3cm}<{\centering}p{0.3cm}<{\centering}p{0.3cm}<{\centering}
p{0.3cm}<{\centering}p{0.5cm}<{\centering}|p{0.76cm}<{\centering}}
\toprule
Methods & \rotatebox{90}{road}&\rotatebox{90}{sidewalk}&\rotatebox{90}{building}&\rotatebox{90}{wall}&\rotatebox{90}{fence}&\rotatebox{90}{pole}&\rotatebox{90}{traffic light}&\rotatebox{90}{traffic sign}&\rotatebox{90}{vegetation}& \rotatebox{90}{terrain} & \rotatebox{90}{sky} & \rotatebox{90}{person} & \rotatebox{90}{rider} & \rotatebox{90}{car} & \rotatebox{90}{truck} & \rotatebox{90}{bus} & \rotatebox{90}{train} &\rotatebox{90}{motorcycle} &\rotatebox{90}{bicycle}&mIoU \\
\midrule
\midrule
FCN~\cite{shelhamer2016fully}&97.4 &78.4 &89.2 &34.9 &44.2 &47.4 &60.1 &65.0 &91.4 &69.3 &93.9 &77.1 &51.4 &92.6& 35.3 &48.6 &46.5 &51.6 &66.8& 65.3 \\
DPN~\cite{liu2015semantic}& 97.5 &78.5& 89.5 &40.4 &45.9 &51.1 &56.8 &65.3& 91.5 &69.4 &94.5& 77.5 &54.2 &92.5 &44.5& 53.4& 49.9 &52.1 &64.8& 66.8 \\
LRR~\cite{ghiasi2016laplacian}& 97.7 &79.9 &90.7 &44.4 &48.6 &58.6 &68.2 &72.0& 92.5 &69.3 &94.7 &81.6 &60.0 &94.0 &43.6 &56.8 &47.2 &54.8 &69.7& 69.7 \\
Deeplabv2~\cite{chen2016deeplab}&97.9 &81.3 &90.3 &48.8 &47.4 &49.6 &57.9 &67.3 &91.9 &69.4 &94.2 &79.8 &59.8 &93.7& 56.5& 67.5 &57.5& 57.7& 68.8&70.4 \\
RefineNet~\cite{Lin_2017_CVPR}& 98.2& 83.3 &91.3 &47.8 &50.4 &56.1 &66.9 &71.3 &92.3 &70.3 &94.8 &80.9 &63.3 &94.5 &64.6& 76.1 &64.3 &62.2 &70.0 & 73.6 \\
DepthSet~\cite{kong2018recurrent}& -& -& -& -& -& -& -&-& -& -& -& -& -& -& -& -& -&-&- & 78.2 \\
ResNet-38~\cite{wu2019wider}& 98.5 &85.7 &93.1& 55.5 &59.1& 67.1& 74.8 &78.7 &93.7 &72.6 &95.5 &86.6 &69.2& 95.7 &64.5 &78.8 &74.1 &69.0& 76.7& 78.4 \\
PSPNet~\cite{Zhao_2017_CVPR}& 98.6& 86.2& 92.9& 50.8& 58.8& 64.0& 75.6& 79.0& 93.4& 72.3& 95.4& 86.5& 71.3& 95.9& 68.2& 79.5& 73.8&69.5&77.2 & 78.4 \\
AAF~\cite{Ke_2018_ECCV}& 98.5& 85.6& 93.0& 53.8& 59.0& 65.9& 75.0& 78.4& 93.7& 72.4&\textbf{95.6}& 86.4& 70.5& 95.9& 73.9& 82.7& 76.9& 68.7& 76.4& 79.1\\
DFN~\cite{yu2018learning}& -& -& -& -& -& -& -&-& -& -& -& -& -& -& -& -& -&-&- & 79.3\\
PSANet~\cite{zhao2018psanet}& -& -& -& -& -& -& -&-& -& -& -& -& -& -& -& -& -&-&-& 80.1 \\
DenseASPP~\cite{yang2018denseaspp}& 98.7 &\textbf{87.1} &93.4 &\textbf{60.7}& 62.7 &65.6 &74.6& 78.5& 93.6 &72.5 &95.4& 86.2& 71.9 &96.0 &\textbf{78.0}& \textbf{90.3} &80.7 &\textbf{69.7} &76.8 & 80.6 \\
\midrule
BFP (ours) & \textbf{98.7}& 87.0& \textbf{93.5}& 59.8& \textbf{63.4}& \textbf{68.9}& \textbf{76.8}& \textbf{80.9}& \textbf{93.7}& \textbf{72.8}& 95.5& \textbf{87.0}& \textbf{72.1} & \textbf{96.0} & 77.6 &89.0 & \textbf{86.9} &69.2&\textbf{77.6}& \textbf{81.4} \\
\bottomrule
\end{tabular}
\end{center}
\caption{Category-wise performance comparison on the Cityscapes test set. Note that the DenseAspp~\cite{yang2018denseaspp} uses stronger backbone DenseNet-161~\cite{Huang_2017_CVPR} than Resnet-101~\cite{he2016deep} we adopt as backbone.}
\label{Table:Cityscapes_results_comparison}
\end{table*}

\textbf{CamVid}~\cite{brostow2008segmentation} is a road scene image segmentation dataset which provides dense pixel-wise annotations for 11 semantic categories.  There are 367 training images, 101 validation images and 233 testing images. The testing results are shown in Table \ref{Table:resultsCamVid}. It shows again that the proposed BFP outperforms previous state-of-the-arts by a large margin.

\begin{figure}[t]
\centering
  \scriptsize{\hspace{4em} Images \hfill \hspace{0.2em} Ours \hfill  Ground Truth \hspace{2.5em} } \\
 \includegraphics[width=0.476\textwidth]{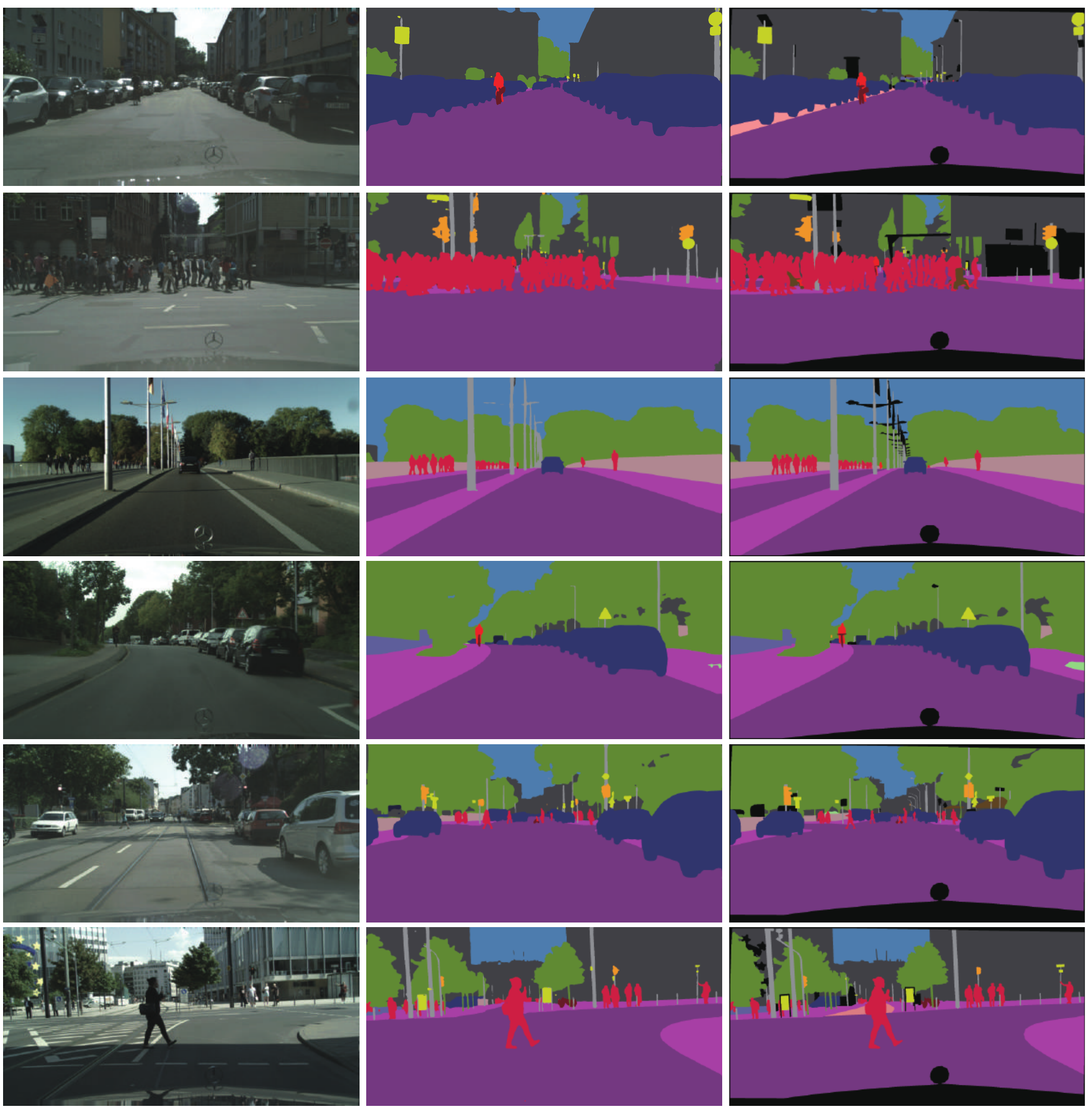}
\caption{Qualitative segmentation examples on Cityscapes.}
\label{Figure:QualitativeResults_Cityscapes}
\end{figure}

\textbf{Cityscapes}~\cite{cordts2016cityscapes} is a recent street scene dataset which contains 5000 high-resolution (1024$\times$2048) images with pixel-level fine annotations. There are 2975 training images, 500 validation images and 1525 testing images. 19 classes (e.g.,~roads, bicycles and cars) are considered for evaluation on the testing sever provided by the organizers. The  Category-wise results are shown in Table \ref{Table:Cityscapes_results_comparison}. Our BFP is only trained on fine annotations, while~\cite{chen2018encoder} also uses coarse annotations for training. Some segmentation examples on Cityscapes are shown in Figure \ref{Figure:QualitativeResults_Cityscapes}.

\vspace{-0.24em}
\section{Conclusion}
\vspace{-0.376em}
In this work, we address the challenging issue of scene segmentation. In order to improve the feature similarity of the same segment while keeping the feature discrimination of different segments, we explore to propagate features throughout the image under the control of inferred boundaries. Towards to this, we first propose to learn the boundary as an additional semantic class to enable the network to be aware of the boundary layout. Then, in order to structurize the image to define the order rules for feature propagation, we propose some unidirectional acyclic graphs (UAGs) to model the function of undirected cyclic graphs (UCGs) in a much more efficient way than DAGs. Based on the proposed UAGs, holistic context is aggregated via harvesting and propagating the local features throughout the whole image efficiently. Finally, we propose a boundary-aware feature propagation (BFP) network to detect and utilize the boundary information for controlling the feature propagation of the UAG-structured image. The proposed BFP is capable of improving the similarity of local features belonging to the same segment region while keeping the discriminative power of features belonging to different segments. We evaluate the proposed boundary-aware feature propagation network on three changeling semantic segmentation datasets, PASCAL-Context, CamVid, and Cityscapes, which show that the proposed BFP achieves new state-of-the-art segmentation performance consistently.

\section*{Acknowledgement}
\begin{spacing}{0.90}
\footnotesize{This research is supported by Singapore Ministry of Education Academic Research Fund AcRF Tier 3 Grant no: MOE2017-T3-1-001, and by BeingTogether Centre, a collaboration between Nanyang Technological University and University of North Carolina at Chapel Hill. BeingTogether Centre is supported by National Research Foundation, Prime Minister Office, Singapore under its International Research Centres in Singapore Funding Initiative.}
\end{spacing}
\hyphenpenalty=100
\tolerance=10
{\small
\bibliographystyle{ieee_fullname}
\bibliography{pami}

\begin{thebibliography}{10}\itemsep=-1pt

\bibitem{arbelaez2011contour}
Pablo Arbelaez, Michael Maire, Charless Fowlkes, and Jitendra Malik.
\newblock Contour detection and hierarchical image segmentation.
\newblock {\em IEEE transactions on pattern analysis and machine intelligence},
  33(5):898--916, 2011.

\bibitem{arnab2016higher}
Anurag Arnab, Sadeep Jayasumana, Shuai Zheng, and Philip~HS Torr.
\newblock Higher order conditional random fields in deep neural networks.
\newblock In {\em European Conference on Computer Vision}. Springer, 2016.

\bibitem{badrinarayanan2017segnet}
Vijay Badrinarayanan, Alex Kendall, and Roberto Cipolla.
\newblock Segnet: A deep convolutional encoder-decoder architecture for image
  segmentation.
\newblock {\em IEEE Transactions on Pattern Analysis \& Machine Intelligence},
  2017.

\bibitem{bansal2016pixelnet}
Aayush Bansal, Xinlei Chen, Bryan Russell, Abhinav Gupta, and Deva Ramanan.
\newblock Pixelnet: Towards a general pixel-level architecture.
\newblock {\em arXiv:1609.06694}, 2016.

\bibitem{bertasius2015deepedge}
Gedas Bertasius, Jianbo Shi, and Lorenzo Torresani.
\newblock Deepedge: A multi-scale bifurcated deep network for top-down contour
  detection.
\newblock In {\em The IEEE Conference on Computer Vision and Pattern
  Recognition}, pages 4380--4389, 2015.

\bibitem{bertasius2015high}
Gedas Bertasius, Jianbo Shi, and Lorenzo Torresani.
\newblock High-for-low and low-for-high: Efficient boundary detection from deep
  object features and its applications to high-level vision.
\newblock In {\em Proceedings of the IEEE International Conference on Computer
  Vision}, pages 504--512, 2015.

\bibitem{Bilinski_2018_CVPR}
Piotr Bilinski and Victor Prisacariu.
\newblock Dense decoder shortcut connections for single-pass semantic
  segmentation.
\newblock In {\em The IEEE Conference on Computer Vision and Pattern
  Recognition (CVPR)}, 2018.

\bibitem{brostow2008segmentation}
Gabriel~J Brostow, Jamie Shotton, Julien Fauqueur, and Roberto Cipolla.
\newblock Segmentation and recognition using structure from motion point
  clouds.
\newblock In {\em European conference on computer vision}, 2008.

\bibitem{Byeon_2015_CVPR}
Wonmin Byeon, Thomas~M. Breuel, Federico Raue, and Marcus Liwicki.
\newblock Scene labeling with lstm recurrent neural networks.
\newblock In {\em The IEEE Conference on Computer Vision and Pattern
  Recognition (CVPR)}, June 2015.

\bibitem{carreira2012semantic}
Joao Carreira, Rui Caseiro, Jorge Batista, and Cristian Sminchisescu.
\newblock Semantic segmentation with second-order pooling.
\newblock {\em Computer Vision--ECCV 2012}, 2012.

\bibitem{chen2016semantic}
Liang-Chieh Chen, Jonathan~T Barron, George Papandreou, Kevin Murphy, and
  Alan~L Yuille.
\newblock Semantic image segmentation with task-specific edge detection using
  cnns and a discriminatively trained domain transform.
\newblock In {\em The IEEE Conference on Computer Vision and Pattern
  Recognition}, pages 4545--4554, 2016.

\bibitem{chen2015semantic}
Liang-Chieh Chen, George Papandreou, Iasonas Kokkinos, Kevin Murphy, and Alan~L
  Yuille.
\newblock Semantic image segmentation with deep convolutional nets and fully
  connected crfs.
\newblock In {\em ICLR}, 2015.

\bibitem{chen2016deeplab}
Liang-Chieh Chen, George Papandreou, Iasonas Kokkinos, Kevin Murphy, and Alan~L
  Yuille.
\newblock Deeplab: Semantic image segmentation with deep convolutional nets,
  atrous convolution, and fully connected crfs.
\newblock {\em arXiv:1606.00915}, 2016.

\bibitem{chen2018encoder}
Liang-Chieh Chen, Yukun Zhu, George Papandreou, Florian Schroff, and Hartwig
  Adam.
\newblock Encoder-decoder with atrous separable convolution for semantic image
  segmentation.
\newblock {\em arXiv:1802.02611}, 2018.

\bibitem{cordts2016cityscapes}
Marius Cordts, Mohamed Omran, Sebastian Ramos, Timo Rehfeld, Markus Enzweiler,
  Rodrigo Benenson, Uwe Franke, Stefan Roth, and Bernt Schiele.
\newblock The cityscapes dataset for semantic urban scene understanding.
\newblock In {\em The IEEE Conference on Computer Vision and Pattern
  Recognition}, 2016.

\bibitem{dai2015boxsup}
Jifeng Dai, Kaiming He, and Jian Sun.
\newblock Boxsup: Exploiting bounding boxes to supervise convolutional networks
  for semantic segmentation.
\newblock In {\em Proceedings of the IEEE International Conference on Computer
  Vision}, 2015.

\bibitem{ding2019semantic}
Henghui Ding, Xudong Jiang, Bing Shuai, Ai~Qun Liu, and Gang Wang.
\newblock Semantic correlation promoted shape-variant context for segmentation.
\newblock In {\em Proceedings of the IEEE Conference on Computer Vision and
  Pattern Recognition (CVPR)}, pages 8885--8894, June 2019.

\bibitem{Ding_2018_CVPR}
Henghui Ding, Xudong Jiang, Bing Shuai, Ai Qun~Liu, and Gang Wang.
\newblock Context contrasted feature and gated multi-scale aggregation for
  scene segmentation.
\newblock In {\em Proceedings of the IEEE Conference on Computer Vision and
  Pattern Recognition (CVPR)}, pages 2393--2402, June 2018.

\bibitem{dollar2013structured}
Piotr Doll{\'a}r and C~Lawrence Zitnick.
\newblock Structured forests for fast edge detection.
\newblock In {\em Proceedings of the IEEE International Conference on Computer
  Vision}, pages 1841--1848, 2013.

\bibitem{dollar2015fast}
Piotr Doll{\'a}r and C~Lawrence Zitnick.
\newblock Fast edge detection using structured forests.
\newblock {\em IEEE transactions on pattern analysis and machine intelligence},
  37(8):1558--1570, 2015.

\bibitem{fu2019dual}
Jun Fu, Jing Liu, Haijie Tian, Yong Li, Yongjun Bao, Zhiwei Fang, and Hanqing
  Lu.
\newblock Dual attention network for scene segmentation.
\newblock In {\em The IEEE Conference on Computer Vision and Pattern
  Recognition}, pages 3146--3154, 2019.

\bibitem{fu2019adaptive}
Jun Fu, Jing Liu, Yuhang Wang, Yong Li, Yongjun Bao, Jinhui Tang, and Hanqing
  Lu.
\newblock Adaptive context network for scene parsing.
\newblock In {\em Proceedings of the IEEE international conference on computer
  vision}, 2019.

\bibitem{ghiasi2016laplacian}
Golnaz Ghiasi and Charless~C Fowlkes.
\newblock Laplacian pyramid reconstruction and refinement for semantic
  segmentation.
\newblock In {\em European Conference on Computer Vision}. Springer, 2016.

\bibitem{gould2009decomposing}
Stephen Gould, Richard Fulton, and Daphne Koller.
\newblock Decomposing a scene into geometric and semantically consistent
  regions.
\newblock In {\em International Conference on Computer Vision}, pages 1--8.
  IEEE, 2009.

\bibitem{gu2018unpaired}
Jiuxiang Gu, Shafiq Joty, Jianfei Cai, and Gang Wang.
\newblock Unpaired image captioning by language pivoting.
\newblock In {\em ECCV}, 2018.

\bibitem{gu2019unpaired}
Jiuxiang Gu, Shafiq Joty, Jianfei Cai, Handong Zhao, Xu Yang, and Gang Wang.
\newblock Unpaired image captioning via scene graph alignments.
\newblock In {\em ICCV}, 2019.

\bibitem{gu2019graph}
Jiuxiang Gu, Handong Zhao, Zhe Lin, Sheng Li, Jianfei Cai, and Mingyang Ling.
\newblock Scene graph generation with external knowledge and image
  reconstruction.
\newblock In {\em CVPR}, 2019.

\bibitem{hayder2017boundary}
Zeeshan Hayder, Xuming He, and Mathieu Salzmann.
\newblock Boundary-aware instance segmentation.
\newblock In {\em The IEEE Conference on Computer Vision and Pattern
  Recognition}, pages 5696--5704, 2017.

\bibitem{he2019dynamic}
Junjun He, Zhongying Deng, and Yu Qiao.
\newblock Dynamic multi-scale filters for semantic segmentation.
\newblock In {\em Proceedings of the International Conference on Computer
  Vision}, 2019.

\bibitem{Hejun_2019_CVPR}
Junjun He, Zhongying Deng, Lei Zhou, Yali Wang, and Yu Qiao.
\newblock Adaptive pyramid context network for semantic segmentation.
\newblock In {\em The IEEE Conference on Computer Vision and Pattern
  Recognition (CVPR)}, June 2019.

\bibitem{he2016deep}
Kaiming He, Xiangyu Zhang, Shaoqing Ren, and Jian Sun.
\newblock Deep residual learning for image recognition.
\newblock In {\em The IEEE Conference on Computer Vision and Pattern
  Recognition}, 2016.

\bibitem{Huang_2017_CVPR}
Gao Huang, Zhuang Liu, Laurens van~der Maaten, and Kilian~Q. Weinberger.
\newblock Densely connected convolutional networks.
\newblock In {\em The IEEE Conference on Computer Vision and Pattern
  Recognition (CVPR)}, 2017.

\bibitem{hwang2015pixel}
Jyh-Jing Hwang and Tyng-Luh Liu.
\newblock Pixel-wise deep learning for contour detection.
\newblock {\em arXiv:1504.01989}, 2015.

\bibitem{islam2017gated}
Md~Amirul Islam, Mrigank Rochan, Neil~DB Bruce, and Yang Wang.
\newblock Gated feedback refinement network for dense image labeling.
\newblock In {\em IEEE Conference on Computer Vision and Pattern Recognition
  (CVPR)}, 2017.

\bibitem{isola2014crisp}
Phillip Isola, Daniel Zoran, Dilip Krishnan, and Edward~H Adelson.
\newblock Crisp boundary detection using pointwise mutual information.
\newblock In {\em European Conference on Computer Vision}, pages 799--814.
  Springer, 2014.

\bibitem{Ke_2018_ECCV}
Tsung-Wei Ke, Jyh-Jing Hwang, Ziwei Liu, and Stella~X. Yu.
\newblock Adaptive affinity fields for semantic segmentation.
\newblock In {\em The European Conference on Computer Vision (ECCV)}, September
  2018.

\bibitem{kohli2009robust}
Pushmeet Kohli, Philip~HS Torr, et~al.
\newblock Robust higher order potentials for enforcing label consistency.
\newblock {\em International Journal of Computer Vision}, 82(3):302--324, 2009.

\bibitem{kokkinos2015pushing}
Iasonas Kokkinos.
\newblock Pushing the boundaries of boundary detection using deep learning.
\newblock {\em arXiv:1511.07386}, 2015.

\bibitem{kong2018recurrent}
Shu Kong and Charless~C Fowlkes.
\newblock Recurrent scene parsing with perspective understanding in the loop.
\newblock In {\em The IEEE Conference on Computer Vision and Pattern
  Recognition}, 2018.

\bibitem{konishi2003statistical}
Scott Konishi, Alan~L. Yuille, James~M. Coughlan, and Song~Chun Zhu.
\newblock Statistical edge detection: Learning and evaluating edge cues.
\newblock {\em IEEE Transactions on Pattern Analysis and Machine Intelligence},
  25(1):57--74, 2003.

\bibitem{krahenbuhl2011efficient}
Philipp Kr{\"a}henb{\"u}hl and Vladlen Koltun.
\newblock Efficient inference in fully connected crfs with gaussian edge
  potentials.
\newblock In {\em Advances in neural information processing systems}, 2011.

\bibitem{krizhevsky2012imagenet}
Alex Krizhevsky, Ilya Sutskever, and Geoffrey~E Hinton.
\newblock Imagenet classification with deep convolutional neural networks.
\newblock In {\em Advances in neural information processing systems}, 2012.

\bibitem{kumar2010efficiently}
M~Pawan Kumar and Daphne Koller.
\newblock Efficiently selecting regions for scene understanding.
\newblock In {\em Computer Vision and Pattern Recognition (CVPR), 2010 IEEE
  Conference on}, pages 3217--3224. IEEE, 2010.

\bibitem{kundu2016feature}
Abhijit Kundu, Vibhav Vineet, and Vladlen Koltun.
\newblock Feature space optimization for semantic video segmentation.
\newblock In {\em The IEEE Conference on Computer Vision and Pattern
  Recognition}, 2016.

\bibitem{larlus2008combining}
Diane Larlus and Fr{\'e}d{\'e}ric Jurie.
\newblock Combining appearance models and markov random fields for category
  level object segmentation.
\newblock In {\em Computer Vision and Pattern Recognition, 2008. CVPR 2008.
  IEEE Conference on}, pages 1--7. IEEE, 2008.

\bibitem{li2017not}
Xiaoxiao Li, Ziwei Liu, Ping Luo, Chen Change~Loy, and Xiaoou Tang.
\newblock Not all pixels are equal: Difficulty-aware semantic segmentation via
  deep layer cascade.
\newblock In {\em The IEEE Conference on Computer Vision and Pattern
  Recognition}, pages 3193--3202, 2017.

\bibitem{liang2016semantic}
Xiaodan Liang, Xiaohui Shen, Donglai Xiang, Jiashi Feng, Liang Lin, and
  Shuicheng Yan.
\newblock Semantic object parsing with local-global long short-term memory.
\newblock In {\em The IEEE Conference on Computer Vision and Pattern
  Recognition}, 2016.

\bibitem{lim2013sketch}
Joseph~J Lim, C~Lawrence Zitnick, and Piotr Doll{\'a}r.
\newblock Sketch tokens: A learned mid-level representation for contour and
  object detection.
\newblock In {\em The IEEE Conference on Computer Vision and Pattern
  Recognition}, pages 3158--3165, 2013.

\bibitem{lin2018multi}
Di Lin, Yuanfeng Ji, Dani Lischinski, Daniel Cohen-Or, and Hui Huang.
\newblock Multi-scale context intertwining for semantic segmentation.
\newblock In {\em Proceedings of the European Conference on Computer Vision
  (ECCV)}, 2018.

\bibitem{Lin_2017_CVPR}
Guosheng Lin, Anton Milan, Chunhua Shen, and Ian Reid.
\newblock Refinenet: Multi-path refinement networks for high-resolution
  semantic segmentation.
\newblock In {\em The IEEE Conference on Computer Vision and Pattern
  Recognition (CVPR)}, 2017.

\bibitem{liu2011sift}
Ce Liu, Jenny Yuen, and Antonio Torralba.
\newblock Sift flow: Dense correspondence across scenes and its applications.
\newblock {\em IEEE transactions on pattern analysis and machine intelligence},
  33(5), 2011.

\bibitem{liu2019feature}
Jun Liu, Henghui Ding, Amir Shahroudy, Ling-Yu Duan, Xudong Jiang, Gang Wang,
  and Alex~Kot Chichung.
\newblock Feature boosting network for 3d pose estimation.
\newblock {\em IEEE transactions on pattern analysis and machine intelligence},
  2019.

\bibitem{liu2015semantic}
Ziwei Liu, Xiaoxiao Li, Ping Luo, Chen-Change Loy, and Xiaoou Tang.
\newblock Semantic image segmentation via deep parsing network.
\newblock In {\em Proceedings of the IEEE International Conference on Computer
  Vision}, 2015.

\bibitem{long2015fully}
Jonathan Long, Evan Shelhamer, and Trevor Darrell.
\newblock Fully convolutional networks for semantic segmentation.
\newblock In {\em The IEEE Conference on Computer Vision and Pattern
  Recognition}, 2015.

\bibitem{mottaghi_cvpr14}
Roozbeh Mottaghi, Xianjie Chen, Xiaobai Liu, Nam-Gyu Cho, Seong-Whan Lee, Sanja
  Fidler, Raquel Urtasun, and Alan Yuille.
\newblock The role of context for object detection and semantic segmentation in
  the wild.
\newblock In {\em IEEE Conference on Computer Vision and Pattern Recognition
  (CVPR)}, 2014.

\bibitem{noh2015learning}
Hyeonwoo Noh, Seunghoon Hong, and Bohyung Han.
\newblock Learning deconvolution network for semantic segmentation.
\newblock In {\em Proceedings of the IEEE International Conference on Computer
  Vision}, 2015.

\bibitem{pascanu2013difficulty}
Razvan Pascanu, Tomas Mikolov, and Yoshua Bengio.
\newblock On the difficulty of training recurrent neural networks.
\newblock In {\em International Conference on Machine Learning}, pages
  1310--1318, 2013.

\bibitem{paszke2016enet}
Adam Paszke, Abhishek Chaurasia, Sangpil Kim, and Eugenio Culurciello.
\newblock Enet: A deep neural network architecture for real-time semantic
  segmentation.
\newblock {\em arXiv:1606.02147}, 2016.

\bibitem{romera2018erfnet}
Eduardo Romera, Jos{\'e}~M Alvarez, Luis~M Bergasa, and Roberto Arroyo.
\newblock Erfnet: Efficient residual factorized convnet for real-time semantic
  segmentation.
\newblock {\em IEEE Transactions on Intelligent Transportation Systems},
  19(1):263--272, 2018.

\bibitem{rota2018place}
Samuel Rota~Bul{\`o}, Lorenzo Porzi, and Peter Kontschieder.
\newblock In-place activated batchnorm for memory-optimized training of dnns.
\newblock In {\em The IEEE Conference on Computer Vision and Pattern
  Recognition}, pages 5639--5647, 2018.

\bibitem{russell2009associative}
Chris Russell, Pushmeet Kohli, Philip~HS Torr, et~al.
\newblock Associative hierarchical crfs for object class image segmentation.
\newblock In {\em Computer Vision, 2009 IEEE 12th International Conference on},
  pages 739--746. IEEE, 2009.

\bibitem{shelhamer2016fully}
Evan Shelhamer, Jonathon Long, and Trevor Darrell.
\newblock Fully convolutional networks for semantic segmentation.
\newblock {\em IEEE transactions on pattern analysis and machine intelligence},
  2016.

\bibitem{shen2017multi}
Wei Shen, Bin Wang, Yuan Jiang, Yan Wang, and Alan Yuille.
\newblock Multi-stage multi-recursive-input fully convolutional networks for
  neuronal boundary detection.
\newblock In {\em Proceedings of the IEEE International Conference on Computer
  Vision}, pages 2391--2400, 2017.

\bibitem{shen2015deepcontour}
Wei Shen, Xinggang Wang, Yan Wang, Xiang Bai, and Zhijiang Zhang.
\newblock Deepcontour: A deep convolutional feature learned by positive-sharing
  loss for contour detection.
\newblock In {\em The IEEE Conference on Computer Vision and Pattern
  Recognition}, pages 3982--3991, 2015.

\bibitem{shuai2019toward}
Bing Shuai, Henghui Ding, Ting Liu, Gang Wang, and Xudong Jiang.
\newblock Toward achieving robust low-level and high-level scene parsing.
\newblock {\em IEEE Transactions on Image Processing}, 28(3):1378--1390, 2019.

\bibitem{shuai2017scene}
Bing Shuai, Zhen Zuo, Bing Wang, and Gang Wang.
\newblock Scene segmentation with dag-recurrent neural networks.
\newblock {\em IEEE transactions on pattern analysis and machine intelligence},
  40(6):1480--1493, 2018.

\bibitem{simonyan2014very}
Karen Simonyan and Andrew Zisserman.
\newblock Very deep convolutional networks for large-scale image recognition.
\newblock {\em arXiv:1409.1556}, 2014.

\bibitem{szegedy2015going}
Christian Szegedy, Wei Liu, Yangqing Jia, Pierre Sermanet, Scott Reed, Dragomir
  Anguelov, Dumitru Erhan, Vincent Vanhoucke, and Andrew Rabinovich.
\newblock Going deeper with convolutions.
\newblock In {\em The IEEE Conference on Computer Vision and Pattern
  Recognition}, 2015.

\bibitem{tighe2013finding}
Joseph Tighe and Svetlana Lazebnik.
\newblock Finding things: Image parsing with regions and per-exemplar
  detectors.
\newblock In {\em The IEEE Conference on Computer Vision and Pattern
  Recognition}, 2013.

\bibitem{wang_iccv2019lfsd}
Tiantian Wang, Yongri Piao, Xiao Li, Lihe Zhang, and Huchuan Lu.
\newblock Deep learning for light field saliency detection.
\newblock In {\em Proceedings of the International Conference on Computer
  Vision}, 2019.

\bibitem{Wang_ECCV2016ksr}
Tiantian Wang, Lihe Zhang, Huchuan Lu, Chong Sun, and Jinqing Qi.
\newblock Kernelized subspace ranking for saliency detection.
\newblock In {\em ECCV}, pages 450--466, 2016.

\bibitem{wang2018detect}
Tiantian Wang, Lihe Zhang, Shuo Wang, Huchuan Lu, Gang Yang, Xiang Ruan, and
  Ali Borji.
\newblock Detect globally, refine locally: A novel approach to saliency
  detection.
\newblock In {\em The IEEE Conference on Computer Vision and Pattern
  Recognition}, pages 3127--3135, 2018.

\bibitem{wu2019wider}
Zifeng Wu, Chunhua Shen, and Anton Van Den~Hengel.
\newblock Wider or deeper: Revisiting the resnet model for visual recognition.
\newblock {\em Pattern Recognition}, 2019.

\bibitem{xie2015holistically}
Saining Xie and Zhuowen Tu.
\newblock Holistically-nested edge detection.
\newblock In {\em Proceedings of the IEEE international conference on computer
  vision}, pages 1395--1403, 2015.

\bibitem{yang2014context}
Jimei Yang, Brian Price, Scott Cohen, and Ming-Hsuan Yang.
\newblock Context driven scene parsing with attention to rare classes.
\newblock In {\em The IEEE Conference on Computer Vision and Pattern
  Recognition}, 2014.

\bibitem{yang2018denseaspp}
Maoke Yang, Kun Yu, Chi Zhang, Zhiwei Li, and Kuiyuan Yang.
\newblock Denseaspp for semantic segmentation in street scenes.
\newblock In {\em The IEEE Conference on Computer Vision and Pattern
  Recognition}, 2018.

\bibitem{Yu_2018_BiSeNet}
Changqian Yu, Jingbo Wang, Chao Peng, Changxin Gao, Gang Yu, and Nong Sang.
\newblock Bisenet: Bilateral segmentation network for real-time semantic
  segmentation.
\newblock In {\em The European Conference on Computer Vision (ECCV)}, 2018.

\bibitem{yu2018learning}
Changqian Yu, Jingbo Wang, Chao Peng, Changxin Gao, Gang Yu, and Nong Sang.
\newblock Learning a discriminative feature network for semantic segmentation.
\newblock {\em arXiv:1804.09337}, 2018.

\bibitem{yu2015multi}
Fisher Yu and Vladlen Koltun.
\newblock Multi-scale context aggregation by dilated convolutions.
\newblock {\em arXiv:1511.07122}, 2015.

\bibitem{ZengY_2018_CVPR}
Yu Zeng, Huchuan Lu, Lihe Zhang, Mengyang Feng, and Ali Borji.
\newblock Learning to promote saliency detectors.
\newblock In {\em The IEEE Conference on Computer Vision and Pattern
  Recognition}, 2018.

\bibitem{zeng2019joint}
Yu Zeng, Yunzhi Zhuge, Huchuan Lu, and Lihe Zhang.
\newblock Joint learning of saliency detection and weakly supervised semantic
  segmentation.
\newblock In {\em Proceedings of the International Conference on Computer
  Vision}, 2019.

\bibitem{Zhang_2018_CVPR}
Hang Zhang, Kristin Dana, Jianping Shi, Zhongyue Zhang, Xiaogang Wang, Ambrish
  Tyagi, and Amit Agrawal.
\newblock Context encoding for semantic segmentation.
\newblock In {\em The IEEE Conference on Computer Vision and Pattern
  Recognition (CVPR)}, 2018.

\bibitem{zhang2018bi}
Lu Zhang, Ju Dai, Huchuan Lu, You He, and Gang Wang.
\newblock A bi-directional message passing model for salient object detection.
\newblock In {\em The IEEE Conference on Computer Vision and Pattern
  Recognition}, 2018.

\bibitem{Zhang2019VOS}
Lu Zhang, Zhe Lin, Jianming Zhang, Huchuan Lu, and You He.
\newblock Fast video object segmentation via dynamic targeting network.
\newblock In {\em Proceedings of the International Conference on Computer
  Vision}, 2019.

\bibitem{Zhao_2018_ICNet}
Hengshuang Zhao, Xiaojuan Qi, Xiaoyong Shen, Jianping Shi, and Jiaya Jia.
\newblock Icnet for real-time semantic segmentation on high-resolution images.
\newblock In {\em The European Conference on Computer Vision (ECCV)}, September
  2018.

\bibitem{Zhao_2017_CVPR}
Hengshuang Zhao, Jianping Shi, Xiaojuan Qi, Xiaogang Wang, and Jiaya Jia.
\newblock Pyramid scene parsing network.
\newblock In {\em The IEEE Conference on Computer Vision and Pattern
  Recognition (CVPR)}, 2017.

\bibitem{zhao2018psanet}
Hengshuang Zhao, Yi Zhang, Shu Liu, Jianping Shi, Chen~Change Loy, Dahua Lin,
  and Jiaya Jia.
\newblock {PSANet}: Point-wise spatial attention network for scene parsing.
\newblock In {\em ECCV}, 2018.

\end{thebibliography}
}
\end{document}